\pdfoutput=1

\documentclass[11pt]{article}
\usepackage[T1]{fontenc}
\usepackage[utf8]{inputenc}
\usepackage{listings}
\usepackage{xcolor}
\usepackage{longtable}
\usepackage{inconsolata}
\usepackage{microtype}

\usepackage{acl2025}

\usepackage{times}
\usepackage{latexsym}

\usepackage{float}
\restylefloat{table}

\usepackage[T1]{fontenc} 

\usepackage[utf8]{inputenc}

\usepackage{color-edits}
\addauthor{gn}{magenta}
\addauthor{byl}{purple}

\usepackage{microtype}
\usepackage{graphicx}
\graphicspath{{image/}}
\usepackage{pdfpages}
\usepackage{url}
\usepackage{hyperref}
\usepackage{xcolor}
\usepackage{epsfig}
\usepackage{adjustbox}
\usepackage{amsfonts}
\usepackage{amsmath}
\usepackage{amssymb}
\usepackage{booktabs} 
\usepackage{comment}
\usepackage{multirow}
\usepackage{caption}
\usepackage{subcaption}
\usepackage{textcomp}
\usepackage{comment}
\usepackage{relsize}
\usepackage{stmaryrd}
\usepackage{bbm}
\usepackage{rotating}
\usepackage{arydshln}
\usepackage{pifont}
\usepackage{xcolor}
\usepackage{colortbl}
\usepackage{tabularray}
\usepackage{graphicx}
\usepackage{pbox}
\usepackage[most]{tcolorbox}
\usepackage{colortbl}
\usepackage{enumitem}
\newtcolorbox{mybox}[2][]{
    colback=white,
    colframe=green!45,
    fonttitle=\bfseries,
    coltitle=black,
    sharp corners,
    title=#2,
    #1
}

\usepackage{color-edits}
\addauthor{gn}{magenta}

\title{Evaluating Language Models as Synthetic Data Generators}

\author{Seungone Kim$^{1}$ \quad Juyoung Suk$^{2}$ \quad Xiang Yue$^{1}$ \quad \textbf{Vijay Viswanathan}$^{1}$\\ \textbf{Seongyun Lee}$^{2}$ \quad \textbf{Yizhong Wang}$^{3}$ \quad \textbf{Kiril Gashteovski}$^{4,5}$ \quad \textbf{Carolin Lawrence}$^{4}$\\ \textbf{Sean Welleck}$^{1}$ \quad \textbf{Graham Neubig}$^{1}$\\ \\ Carnegie Mellon University$^{1}$ \qquad KAIST AI$^{2}$ \qquad University of Washington$^{3}$\\ NEC Laboratories Europe$^{4}$ \qquad Ss. Cyril and Methodius University of Skopje$^{5}$\\
\texttt{\{seungone, wellecks, gneubig\}@cmu.edu}}

\begin{document}
\maketitle
\begin{abstract}

Given the increasing use of synthetic data in language model (LM) post-training, an LM's ability to generate high-quality data has become nearly as crucial as its ability to solve problems directly. While prior works have focused on developing effective data generation methods, they lack systematic comparison of different \emph{LMs as data generators} in a unified setting. To address this gap, we propose \textsc{AgoraBench}, a benchmark that provides standardized settings and metrics to evaluate LMs' data generation abilities. Through synthesizing 1.26 million training instances using 6 LMs and training 99 student models, we uncover key insights about LMs' data generation capabilities. First, we observe that LMs exhibit distinct strengths. For instance, GPT-4o excels at generating new problems, while Claude-3.5-Sonnet performs better at enhancing existing ones. Furthermore, our analysis reveals that an LM's data generation ability doesn't necessarily correlate with its problem-solving ability. Instead, multiple intrinsic features of data quality—including response quality, perplexity, and instruction difficulty—collectively serve as better indicators. Finally, we demonstrate that strategic choices in output format and cost-conscious model selection significantly impact data generation effectiveness. Our code, checkpoints, and data are all publicly available at \href{https://github.com/neulab/data-agora}{https://github.com/neulab/data-agora}.
\end{abstract}


\section{Introduction}



Post-training language models (LMs) on synthetic data is a promising approach for improving their ability to solve a wide range of tasks~\citep{wang2023self,honovich2023unnatural,alpaca,liu2024best}. While acquiring data through manual annotation continues to play an important role, synthetic data generation offers a scalable complement to human labeling~\citep{viswanathan2023prompt2model,kim2023cotever}. Hence, numerous works have proposed novel methods to effectively generate high-quality synthetic data~\citep{xu2024wizardlm,gunasekar2023textbooks,yue2023mammoth,yue2024mammoth2}.


As multiple proprietary LMs with comparable performance emerge and open-source LMs steadily catch up~\citep{hurst2024gpt,anthropic2024claude,llama3,qwen2.5}, measuring each LM's data generation capabilities has become as crucial as developing new data generation methods. Moreover, companies providing proprietary LMs have begun promoting the use of their latest models for generating synthetic data~\citep{nvidia2024synthetic}.
Carefully comparing data generation ability across LMs helps validate these claims and allows practitioners to wisely choose models for data synthesis.

\begin{figure*}[h] 
\centering
\includegraphics[width=0.8\textwidth]{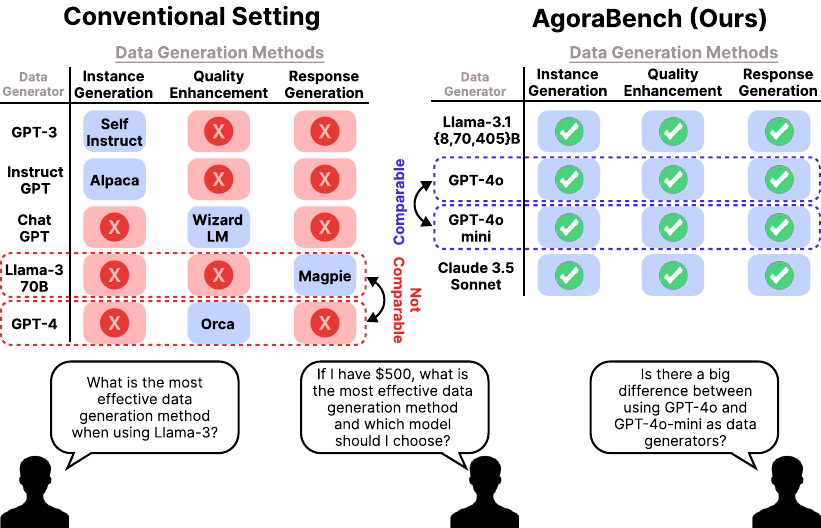}
\caption{\textbf{Illustration of the motivation of \textsc{AgoraBench}:} Prior works focused on developing new methods to generate synthetic data. In contrast, our work focuses on systematically comparing different LMs as data generators based on existing data generation methods. Further explanation of data generation methods are covered in Section~\ref{sec:setting}.}
\vspace{-3mm}
\label{fig:agorabench_illustration}
\end{figure*}

To systematically compare LMs' capabilities as data generators, a \textit{unified experimental setting} is needed, where only the data generator varies, while other components remain fixed. However, as shown in Figure~\ref{fig:agorabench_illustration}, prior works have focused more on demonstrating the effectiveness of their data generation method, leading to various experimental settings that make such comparison challenging. For instance, Self-Instruct~\citep{wang2023self}, Alpaca~\citep{alpaca}, WizardLM~\citep{xu2024wizardlm} and Orca~\citep{mukherjee2023orca} varied in their choice of LMs for data generation, quantity of synthetic training data, base models used for training, and benchmarks for evaluating the model trained on the synthetic dataset. These heterogeneous settings make it difficult to isolate and measure LMs' data generation capabilities, highlighting the need for controlled settings.

To this end, we propose \textbf{\textsc{AgoraBench}}, a benchmark for evaluating LMs' data generation capabilities across nine settings, combining three domains (math, instruction-following, code) with three data generation methods (instance generation, response generation, quality enhancement). Within each setting, all variables except the data generator are controlled: the same meta-prompt and seed dataset are used, with each LM generating an identical number of training instances. Llama-3.1-8B is trained on each synthetic dataset and evaluated on a fixed set of benchmarks spanning different capabilities: mathematics, coding, and general instruction-following. To evaluate the quality of synthetic data, we define a metric called \textbf{Performance Gap Recovered} (PGR), which measures the relative improvement of the model trained on the data (denoted as `student model') over its base model. Based on this setting, we assess six LMs as data generators: GPT-4o, GPT-4o-mini, Claude-3.5-Sonnet, and Llama-3.1-Instruct (8, 70, 405B).%
\footnote{
\citet{xu2024stronger}, a contemporaneous work with ours, also measured various LMs' data generation capabilities.
In contrast to our work, they only examine the ``response generation'' setting, whereas we measure three data generation settings and also do a number of additional analyses on the relationship between the intrinsic quality of data and PGR.
}

Our analysis reveals distinct strengths among different LMs in various kinds of data generation methods. For example, GPT-4o demonstrates superior performance in generating new instances (+ 46.75\%), outperforming both Claude-3.5-Sonnet (+ 24.14\%) and Llama-3.1-405B-Instruct (+ 10.10\%). On the other hand, Claude-3.5-Sonnet excels at refining existing instances (+ 17.89\%), surpassing both GPT-4o (+ 6.69\%) and GPT-4o-mini (+ 5.49\%). These findings demonstrate how \textsc{AgoraBench} can guide practitioners in selecting appropriate LMs for their specific needs.

Unexpectedly, we also find that LMs with weaker problem-solving ability sometimes outperform stronger ones in data generation—for example, Claude-3.5-Sonnet (+ 23.43\%) is less effective than Llama-3.1-8B-Instruct (+ 55.69\%) at generating new instances in the code domain. Based on these findings, we investigate whether an LM's data generation ability can be predicted by its problem-solving ability alone. Our analysis reveals no strong correlation between the two capabilities. Instead, multiple intrinsic features of data quality—including instruction difficulty, response quality, and response perplexity—collectively influence the student model's improvement. Furthermore, we demonstrate that the top-5 principal components extracted from intrinsic measurements can explain 93.4\% of the variance in the PGR values.

\begin{figure*}[h] 
\centering
\includegraphics[width=\textwidth]{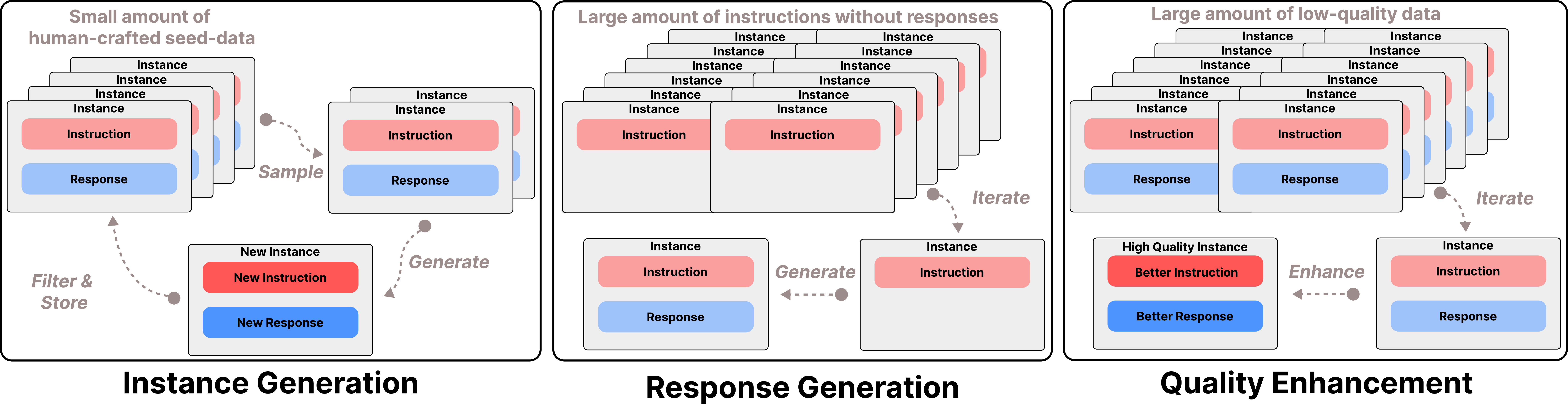}
\vspace{-6mm}
\caption{\textbf{\textsc{AgoraBench} tests three data generation methods}: generating new instruction and response pairs (left), generating responses (middle), and enhancing the quality of the instruction and/or the response (right).}
\label{fig:main}
\end{figure*}

\section{Preliminaries: Measuring Data Generation Capabilities of LMs}\label{sec:setting}



\paragraph{Notations.}Given seed data ${D}_{seed}$ and a prompt describing the kind of data generation to perform (referred to as a `meta-prompt') $M$, a data generator $G$ generates
\begin{equation}
    D_{G} = G({D}_{seed}, M),
\end{equation}
where ${D}_{seed}$ and $D_{G}$ can both be expressed as $\{( I_i, R_i) \mid i = 1, \dots, n\}$ with $I$ denoting an instruction, $R$ denoting a corresponding response, and $n$ denoting the size of the data.

\paragraph{Data Generation Methods.} As shown in Figure~\ref{fig:main}, among the various methods for generating data, most can be grouped into three categories: instance generation, response generation, and quality enhancement. These methods work as follows:
\begin{itemize}[leftmargin=*]
    \item \textbf{Instance Generation}: Given a small seed dataset ${D}_{seed} = \{( I_i, R_i) \mid i = 1, \dots, m\}$, a few instances are randomly sampled from  ${D}_{seed}$ and used as in-context demonstrations, resulting in the generation of new instances~\citep{honovich2023unnatural,wang2023self}. This process is performed iteratively until $D_G = \{( I_i, R_i) \mid i = 1, \dots, n\}$ is constructed where $m << n$. Note that the generated instances could also optionally be used as demonstrations as well.
    \item \textbf{Response Generation}: A large set of instructions $D_I = \{(I_i) \mid i = 1, \dots, n\}$ is given, and $G$ iterates through each instruction $I_i$ to generate a corresponding response $R_i$ ~\citep{xu2024magpie}. 
    \item \textbf{Quality Enhancement}: A large set of instructions and responses $D' = \{( I'_i, R'_i) \mid i = 1, \dots, n\}$ is given. $G$ iterates through each instance to refine $I'_i$ and/or $R'_i$, such as by explicitly prompting $G$ to ``make either/both $I'_i$, $R'_i$ of higher quality'' (\textit{e.g.,} making the instruction more difficult or of higher educational value)~\citep{xu2024wizardlm,yue2024mammoth2}.
\end{itemize}

\begin{table*}[t]
\caption{\textbf{\textsc{AgoraBench} Settings:} For each of the nine settings, an LM being evaluated generates 10K instances with the same meta-prompt and seed data. Note that the seed dataset is also used for training in instance generation.}
\vspace{3mm}
    \centering
    \resizebox{\linewidth}{!}{
        \begin{tabular}{@{}ccccccccccc@{}}
            \toprule
            \textbf{Domain}& \textbf{Data Generation Method} & \textbf{Seed Data} & \textbf{Seed Data Size} &  \textbf{Benchmark}\\
            \midrule
             \multicolumn{1}{c}{\multirow{3}{*}{\textbf{Math}}} &\textbf{Instance Generation} & GSM8K, MATH (train set) & 14,856 &  GSM8K, MATH (test set)\\
            &\textbf{Response Generation} & Magpie-Reasoning (math) &  10,000 &GSM8K, MATH (test set)\\
            &\textbf{Quality Enhancement} & WebInstruct (math) &  10,000 &GSM8K, MATH (test set)\\
            \midrule
            \multicolumn{1}{c}{\multirow{3}{*}{\textbf{Code}}} &\textbf{Instance Generation} & MBPP (train set), xP3x & 874 & MBPP, HumanEval (test set)\\
            &\textbf{Response Generation} & Magpie-Reasoning (code) &  10,000 &MBPP, HumanEval (test set)\\
            &\textbf{Quality Enhancement} & CoNaLa & 10,000 & MBPP, HumanEval (test set)\\
            \midrule
            \multicolumn{1}{c}{\multirow{3}{*}{\textbf{Inst. Follow}}} &\textbf{Instance Generation} & LIMA & 503 & AlpacaEval 2.0, Arena-Hard\\
            &\textbf{Response Generation} & Magpie-Pro &  10,000 &AlpacaEval 2.0, Arena-Hard\\
            &\textbf{Quality Enhancement} & WebInstruct (code) & 10,000 & AlpacaEval 2.0, Arena-Hard\\
            \bottomrule
        \end{tabular}
    }
    \label{tab:eval-sets}
\end{table*}

\begin{figure}[t] 
\centering
\includegraphics[width=0.48\textwidth]{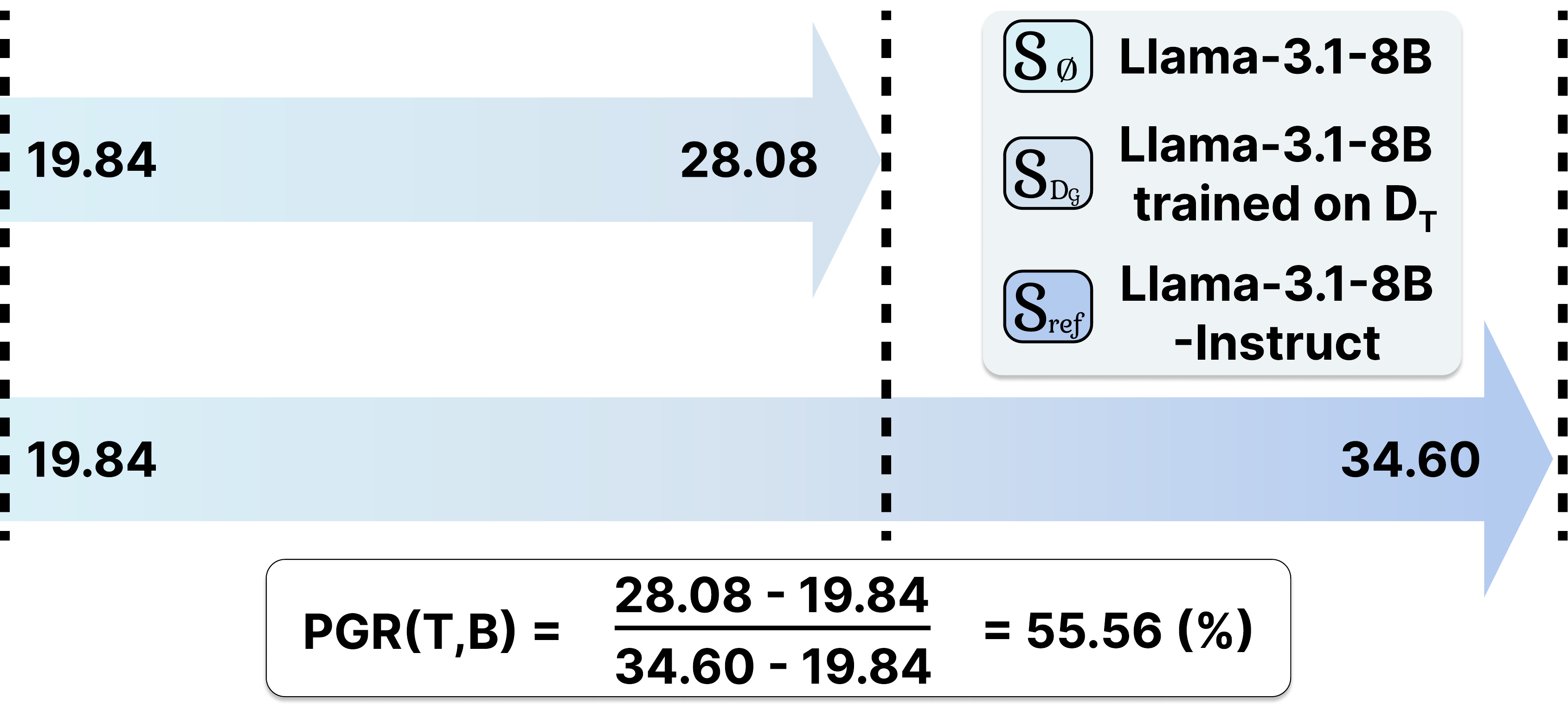}

\caption{\textbf{Illustration of Performance Gap Recovered metric:} The performance gap recovered metric captures the relative improvement of $S_{D_G}$ with respect to $S_{ref}$ where $S_{D_G}$ and $S_{ref}$ is both trained from $S_{\text{\O}}$.}
\label{fig:pgr_example}
\end{figure}

\paragraph{Metric.} An LM's data generation ability can be measured by evaluating the performance improvement of a student model trained on the teacher-generated data. Specifically, we propose a metric, \textbf{Performance Gap Recovered (PGR)}, that measures the improvement on a benchmark $B$ relative to a reference model,
\begin{equation}
PGR(G,B) = \frac{{\text{score}}_{B}({S}_{D_G}) - {\text{score}}_{B}({S}_{\text{\O}})}{{\text{score}}_{B}({S}_{ref}) - {\text{score}}_{B}({S}_{\text{\text{\O}}})} \times 100
\end{equation}

\noindent where ${S}_{\text{\O}}$ denotes a pre-trained LM, ${S}_{D_G}$ denotes ${S}_{\text{\O}}$ trained on $D_G$, ${S}_{ref}$ denotes a reference model that shares the same pre-trained model ${S}_{\text{\O}}$ as a base model, and ${\text{score}}_{B}(\cdot)$ denotes the score on benchmark $B$. In our experiments, we use Llama-3.1-8B as ${S}_{\text{\O}}$ and Llama-3.1-8B-Instruct as ${S}_{ref}$.%
\footnote{
Note that when measuring ${\text{score}}_{B}({S}_{\text{\O}})$, ${S}_{\text{\O}}$ can not solve tasks with zero-shot prompting, so we evaluate their performance with few-shot prompting.
In contrast, ${S}_{D_G}$ and ${S}_{ref}$ are evaluated with zero-shot prompting.
}

Intuitively, as illustrated in Figure~\ref{fig:pgr_example}, by using Llama-3.1-8B as ${S}_{\text{\O}}$ and Llama-3.1-8B-Instruct as ${S}_{ref}$, the PGR value represents how much performance was recovered compared to the post-training process for Llama-3.1-8B-Instruct, which was reportedly extensive, training on 10M+ examples of human-curated data~\citep{llama3}. For example, a PGR value of 50\% indicates that ${S}_{D_G}$ has recovered 50\% of the improvement achieved by ${S}_{ref}$ relative to ${S}_{\text{\O}}$. A value above 100\% indicates ${S}_{D_G}$ outperforms ${S}_{ref}$, while a negative value indicates that training on $D_G$ degraded performance on $B$ compared to few-shot prompting ${S}_{\text{\O}}$.

\paragraph{PGR vs Intrinsic Metrics.} Note that prior works have mostly employed intrinsic metrics to measure data quality, such as response quality, instruction difficulty, and instance diversity. However, these metrics do not directly measure their impact on the ultimate goal for data generation: improving the student model. By introducing PGR (an extrinsic metric), we provide a way to evaluate this improvement directly across data generators. Without this direct measurement, it becomes challenging to assess how much of a student model's improvement stems from the data generator itself versus other contributing factors. Hence, this makes PGR an important complement to intrinsic metrics rather than a replacement for them. We further compare the relationship of how we could use intrinsic metrics to estimate the PGR values in Section~\ref{subsec:predicting}.

\paragraph{Training Student Models.} When training the student model ($S_{\text{\O}}$), we employ supervised fine-tuning (SFT), computing the loss only on response tokens. We directly use the generated data $D_G$ without filtering and do not consider other post-training methods, as our goal is to evaluate the raw data generation capabilities of an LM ($G$) in the most straightforward setting, not to maximize ${S}_{D_G}$'s benchmark performance. The hyper-parameters for training are detailed in Appendix~\ref{appendix:hyper_params}.


\begin{table*}[!t]
\caption{\textbf{\textsc{AgoraBench Results}: How much performance could you recover by generating 10K instances with your LLM, compared to Meta’s post-training process for training Llama-3.1-8B-Instruct from Llama-3.1-8B?} The best comparable performances (\%) are \textbf{bolded}, and the second-best performances (\%) are \underline{underlined}. Note that the Llama models are instruction-tuned versions and that `Inst.' denotes instruction-following.}
\centering
\resizebox{\linewidth}{!}{
\begin{tabular}{lcccccccccccc}
\toprule
\multicolumn{1}{c}{\multirow{2}{*}{\textbf{Data Generator}}} & \multicolumn{4}{c}{\textbf{Instance Generation}} &\multicolumn{4}{c}{\textbf{Response Generation}} &\multicolumn{4}{c}{\textbf{Quality Enhancement}} \\
\cmidrule(lr){2-5} \cmidrule(lr){6-9} \cmidrule(lr){10-13}
              & Math & Code & Inst. & Avg & Math & Code & Inst. & Avg & Math & Code & Inst. & Avg \\
\cmidrule(lr){1-1} \cmidrule(lr){2-2} \cmidrule(lr){3-3} \cmidrule(lr){4-4} \cmidrule(lr){5-5} \cmidrule(lr){6-6} \cmidrule(lr){7-7} \cmidrule(lr){8-8} \cmidrule(lr){9-9} \cmidrule(lr){10-10} \cmidrule(lr){11-11} \cmidrule(lr){12-12} \cmidrule(lr){13-13}
GPT-4o & \textbf{20.6} & \textbf{73.6} & \textbf{46.1} & \textbf{46.8} & \underline{46.7} & 28.5 & \textbf{30.3} & \textbf{35.2} & \textbf{21.9} & -8.8 & 7.1 & \underline{6.7} \\
GPT-4o-mini & \underline{16.1} & 41.9 & 18.0 & \underline{25.3} & \textbf{48.1} & 18.9 & \underline{13.7} & 26.9 & \underline{17.8} & -11.2 & \underline{9.9} & 5.5\\
Claude-3.5-Sonnet & 8.9 & 23.4 & \underline{40.1} & 24.1 & 29.0 & \textbf{44.5} & 12.7 & \underline{28.8} & 15.7 & \textbf{16.1} & \textbf{21.8} & \textbf{17.9}\\
Llama-3.1-405B  & 10.4 & 12.6 & 7.4 & 10.1 & 31.7 & 35.4 & 4.9 & 24.0 & -11.8 & 7.5 & 3.6 & -0.2\\
Llama-3.1-70B  & 9.6 & \underline{58.7} & 6.5 & 24.9 & 23.0 & \underline{37.1} & 4.5 & 21.5 & -21.8 & 6.9 & 2.7 & -4.1\\
Llama-3.1-8B  & 6.5 & 55.7 & 6.2 & 22.8 & 27.6 & 25.8 & 5.0 & 19.4 & -1.7 & \underline{15.4} & 3.0 & 5.6\\
\bottomrule
\end{tabular}
}
\label{table:main_results}
\end{table*}

\section{Experimental Setting of \textsc{AgoraBench}}\label{sec:exp_setting}



Among various choices, \textsc{AgoraBench} focuses on three core capabilities that are considered crucial for LMs: instruction following, mathematical reasoning, and coding~\citep{chang2024survey,guo2023evaluating,hurst2024gpt,anthropic2024claude}. The overall experimental setting of \textsc{AgoraBench} including the domains, seed datasets, and benchmarks for each setting is listed in Table~\ref{tab:eval-sets}.

\paragraph{Domains.} \textsc{AgoraBench} encompasses three domains: math, code, and instruction following. Evaluating three data generation methods across each domain results in nine distinct settings, each with a dedicated seed dataset ($D_{seed}$) and benchmark ($B$). For each setting, the LM employed as the data generator produces 10K training instances. 

Then, the student model is trained using data from a single domain to isolate the effect of generated data quality, as cross-domain training could introduce confounding factors through positive or negative transfer (\textit{e.g.}, training on code data improve math~\citep{dong2023abilities,zhang2024unveiling}).

\paragraph{Seed Datasets.} For each setting, we select seed datasets ($D_{seed}$) based on different assumptions:

\begin{itemize}[leftmargin=*]

\item For \textbf{instance generation}, since we expand a small amount of high-quality data into a larger volume, our approach is premised on using high-quality, human-crafted data as seed data. Hence, we use the train subsets of GSM8K~\citep{cobbe2021training} and MATH~\citep{hendrycks2measuring} for math, MBPP~\citep{austin2021program} and xP3x~\citep{muennighoff2023crosslingual} for code, and LIMA~\citep{zhou2024lima} for instruction following. We exclude instances that exceed 4,096 tokens based on the Llama-3 tokenizer, resulting in 14,856, 874, and 503 seed instances for each of the math, code, and instruction following domains, respectively.

\item For \textbf{response generation}, we simulate how different data generators can attach responses to a fixed set of instructions to ultimately create better quality data. While we could take arbitrary data and discard their responses for experiments, we utilize the Magpie dataset because \citet{xu2024magpie}'s setting closely matches our setting - they first extract instructions by prompting LMs with empty chat templates and then generate responses using two different types of LMs (Llama-3-70B-Instruct and Qwen-2-72B-Instruct).
In our experiments, we sample 10K instances from the Magpie dataset \citep{xu2024magpie} for the instruction following domain and also 10K instances from the Magpie-Reasoning dataset for both math and code domains. 

\item For \textbf{quality enhancement}, we test scenarios where complete instances of instructions and responses already exist, but their quality needs improvement before being used for post-training - either because the instructions are too simple or the responses are not sufficiently detailed. We sample 10K instances from WebInstruct (Q-A pairs from the web requiring refinement; see \citet{yue2024mammoth2}) for instruction following and math domains. Note that WebInstruct does not contain domain labels, hence we prompt GPT-4o-mini-2024-07-18 to prepare a separate $D_{seed}$ (further details are in Appendix~\ref{appendix:webinstruct}). For the code domain, we use CoNaLa, which contains simple instructions paired with 1-3 line code snippets from StackOverflow~\citep{yin2018mining}.

\end{itemize}

\paragraph{Benchmarks.} We evaluate a student model (${S}_{D_G}$)'s performance using two representative benchmarks for each domain. For math, we use the test subsets of GSM8K~\citep{cobbe2021training} and MATH~\citep{hendrycks2measuring}. For code, we use the test set of MBPP~\citep{austin2021program} and HumanEval~\citep{chen2021evaluating}. For instruction following, we evaluate on AlpacaEval-2.0~\citep{dubois2024length} and Arena-Hard~\citep{li2024crowdsourced}.

\section{Experimental Results of \textsc{AgoraBench}}\label{sec:extrinsic_eval}



We compare 6 LMs as data generators ($G$), namely GPT-4o-2024-08-06~\citep{hurst2024gpt}, GPT-4o-mini-2024-07-18, Claude-3.5-Sonnet-2024-06-20~\citep{anthropic2024claude}, Llama-3.1-405B-Instruct, Llama-3.1-70B-Instruct, and Llama-3.1-8B-Instruct~\citep{dubey2024llama}. Also, we use Llama-3.1-8B as the student model (${S}_{\text{\O}}$). The \textsc{AgoraBench} results are listed in Table~\ref{table:main_results}.


\paragraph{GPT-4o is the overall most performant data generator:} Out of the nine experimental settings, GPT-4o achieves the highest PGR scores in five settings. Its performance is particularly notable in instance generation, where it outperforms other LMs as a data generator across all three domains (math at 20.6\%, code at 73.6\%, instruction following at 46.1\%, and total average at 46.8\%), while also achieving the highest average PGR score in response generation (35.2\%). 

\paragraph{Claude-3.5-Sonnet proves particularly effective for quality enhancement:} Claude-3.5-Sonnet particularly demonstrates strong performance in quality enhancement, achieving the highest PGR scores in two out of three domains (code at 21.8\%, instruction following at 17.9\%, and total average at 17.9\%). Additionally, it obtains the best PGR score at response generation in the code domain (44.5\%), bringing its total number of top performances to three out of nine settings.

\begin{table}[!t]
\caption{\textbf{Comparison of API costs, problem-solving ability, and data generation ability:} Our findings reveal that neither the strength nor the cost of an LM guarantees its effectiveness as a data generator. Note that the Llama models are instruction-tuned versions, the specific results of the LMs on each benchmark (averaged as `Problem Solving average') is in Appendix~\ref{appendix:teacher_model_performance}, and the AgoraBench results are averaged from Table~\ref{table:main_results}.}
\centering
\resizebox{\linewidth}{!}{
\begin{tabular}{l|cc|c|c}
\toprule
\multicolumn{1}{c}{\multirow{4}{*}{\textbf{Data Generator}}}  & \multicolumn{2}{c}{\multirow{2}{*}{\textbf{API Cost}}} & \multicolumn{1}{c}{\multirow{1}{*}{\textbf{Prob.}}} &\multicolumn{1}{c}{\multirow{1}{*}{\textbf{Data}}}\\
& & & \textbf{Solv.} & \textbf{Gen.}\\
\cmidrule(lr){2-3} \cmidrule(lr){4-4} \cmidrule(lr){5-5}
              & \multicolumn{1}{c}{\multirow{2}{*}{\textbf{Input}}} & \multicolumn{1}{c}{\multirow{2}{*}{\textbf{Output}}} & \multicolumn{1}{c}{\multirow{2}{*}{\textbf{Avg}}} & \textbf{Agora} \\
              && &&\textbf{Bench} \\
\cmidrule(lr){1-1} \cmidrule(lr){2-2} \cmidrule(lr){3-3} \cmidrule(lr){4-4} \cmidrule(lr){5-5} 
GPT-4o & \$2.50 & \$10.00 & 80.9 & 29.5\%\\
GPT-4o-mini & \$0.15 & \$0.60 & 75.4 & 19.2\%\\
Claude-3.5-Sonnet & \$3.00 & \$15.00 & 80.5 & 23.6\%\\
Llama-3.1-405B & \$1.79 & \$1.79 & 75.0 & 11.3\%\\
Llama-3.1-70B & \$0.35 & \$0.40 & 69.6 & 14.1\%\\
Llama-3.1-8B & \$0.055 & \$0.055 & 50.2 & 15.9\%\\
\bottomrule
\end{tabular}
}
\label{table:cost_teacher_model}
\end{table}

\begin{figure*}[h] 
\centering
\includegraphics[width=0.9\textwidth]{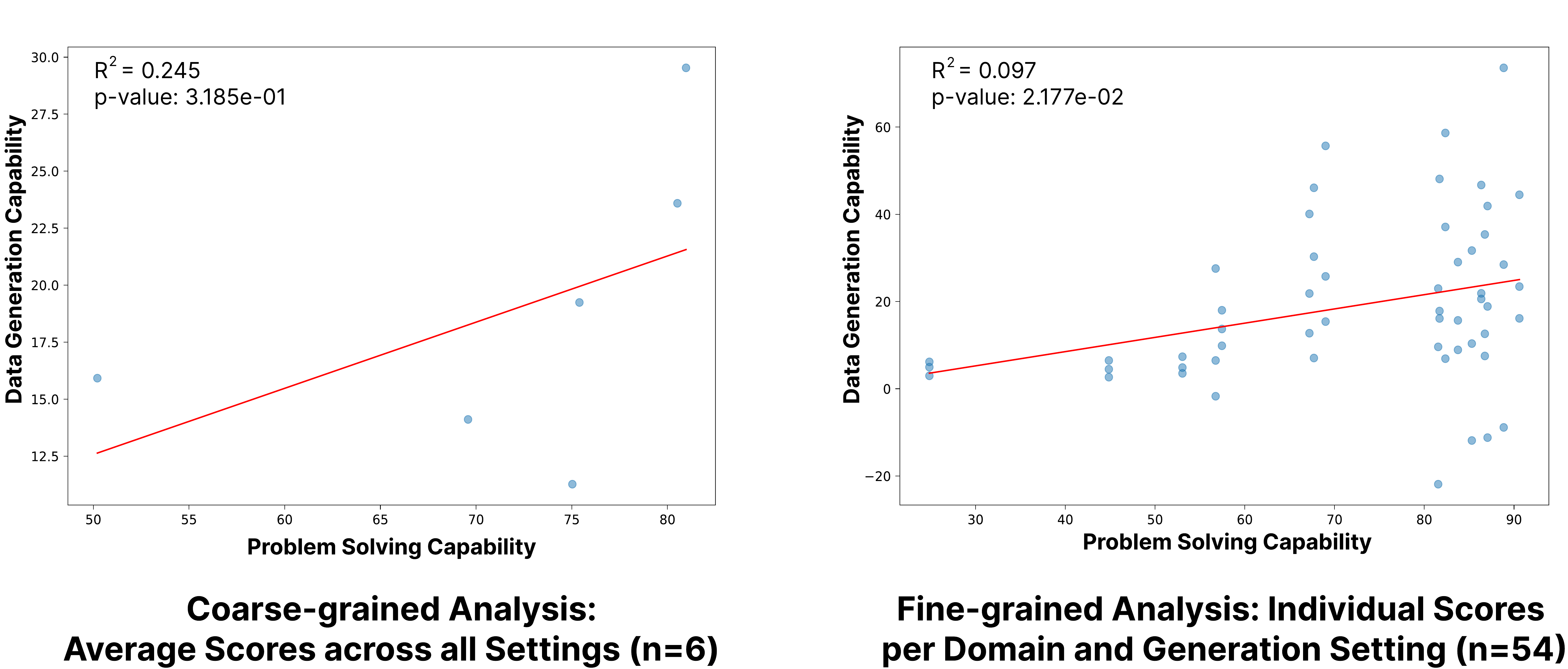}
\caption{\textbf{Problem-solving and data generation capabilities do not strongly correlate: } Linear regression between problem-solving ability and data generation ability scores at multiple granularity levels yields either low $R^2$ values ($R^{2} < 0.1$) or non-significant relationships ($p > 0.05$), which indicates that it is hard to predict data generation capabilities only using problem-solving capabilities.}
\label{fig:linear_regression1}
\end{figure*}

\paragraph{Weaker LMs can outperform Stronger LMs:} We observe cases where LMs with weaker problem-solving abilities achieve higher Performance Gap Recovered (PGR) scores than their stronger counterparts. In the code domain of instance generation, both Claude-3.5-Sonnet (23.4\%) and Llama-3.1-405B-Instruct (12.6\%) are outperformed by Llama-3.1-70B-Instruct (58.7\%) and Llama-3.1-8B-Instruct (55.7\%). Similarly, in the code domain's quality enhancement setting, GPT-4o (-8.8\%) and GPT-4o-mini (-11.2\%) show poorer performance compared to other LMs. 

Interestingly, as shown in Table~\ref{table:cost_teacher_model}, the LMs that performed worse for these cases actually score higher on code benchmarks (MBPP and HumanEval), indicating that they possess stronger problem-solving capabilities. This contradiction suggests that a stronger LM does not necessarily generate better training data. We discuss this phenomenon further in Section~\ref{sec:intrinsic_eval}.


\paragraph{GPT-4o, GPT-4o-mini, and Llama-3.1-8B-Instruct are effective data generators that balance both cost and performance:} Cost is another crucial factor alongside performance when generating large amounts of synthetic data. Table~\ref{table:cost_teacher_model} shows the API costs\footnote{Pricing is based on \href{https://openrouter.ai/}{https://openrouter.ai/}.} benchmark scores (\textit{i.e.,} problem-solving ability) and average performance on \textsc{AgoraBench} (\textit{i.e.,} data generation ability) are listed in Table~\ref{table:cost_teacher_model}. and average performance on \textsc{AgoraBench} for all six LMs. Llama-3.1-8B-Instruct outperforms both Llama-3.1-70B-Instruct and Llama-3.1-405B-Instruct while being 6 to 32.5 times less expensive. Similarly, GPT-4o achieves better performance than Claude-3.5-Sonnet at 1.2 to 1.5 times lower cost. These findings suggest that using more expensive LMs does not necessarily guarantee better data generation, highlighting the importance of careful model selection based on specific tasks or domains of interest.


\section{What makes an effective data generator?}\label{sec:intrinsic_eval}


\begin{figure*}[h] 
\centering
\includegraphics[width=\textwidth]{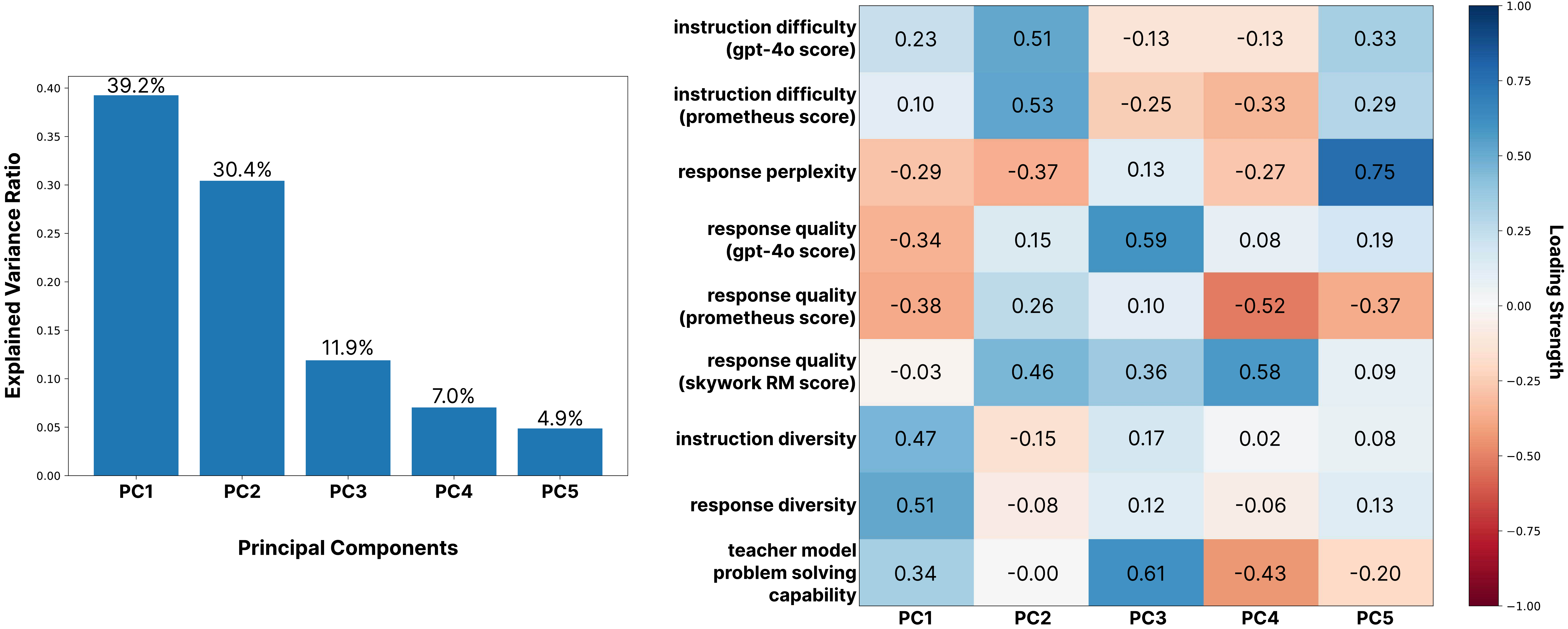}
\caption{Through a PCA analysis on multiple intrinsic evaluation metrics, we find that there exists \textbf{interpretable} low-dimension principal components that \textbf{explain} the variance of data generation capabilities up to 93.4\%.}
\label{fig:pca}
\end{figure*}

In the previous section, we observed an unexpected finding: LMs with weaker problem-solving ability sometimes outperform stronger LMs when generating the same amount of synthetic data under identical conditions. For a better understanding, we examine whether there exists a strong correlation between problem-solving ability and data generation ability (Section~\ref{subsec:best}). Then we investigate whether we can predict the degree of improvement in the performance of the student model by analyzing the data generated by each LM (Section~\ref{subsec:predicting}).

\subsection{Is the best solver necessarily the best generator?}\label{subsec:best}

To examine the relationship between data generation and problem-solving capabilities, we performed linear regression analyses comparing two metrics: average performance on multiple benchmarks (GSM8K, MATH, MBPP, HumanEval, AlpacaEval-2.0, Arena-Hard) and scores from \textsc{AgoraBench}. We conduct this analysis at two levels of granularity. The first analysis (coarse-grained) uses the overall average \textsc{AgoraBench} score across all domains and data generation settings. The second analysis (fine-grained) examines individual scores from different domains and data generation settings in \textsc{AgoraBench} separately.

The results shown in Figure~\ref{fig:linear_regression1} reveal no strong linear correlation between problem-solving capabilities (benchmark scores) and data generation capabilities (\textsc{AgoraBench} PGR scores) at either granularity level. This finding suggests that an LM's performance on traditional benchmarks may not predict its effectiveness as a data generator.

\subsection{Can we predict the student model's improvement by looking into the data?}\label{subsec:predicting}

Given that problem-solving ability does not directly predict data generation ability, we explore what other characteristics might define effective data generators. We hypothesize that good data capable of substantially improving student models share extractable features that can be identified by analyzing their intrinsic properties.
This understanding is crucial as it informs us of what properties the data from a good data generator might possess.
Inspired by \citet{liumakes2023}, we conduct an intrinsic evaluation by analyzing various properties of the generated data $D_G$.

\paragraph{Intrinsic Evaluation Metrics}
We evaluate (1) the complexity of the instruction $I_i$ (2) the quality of response $R_i$, (3) the perplexity of $R_i$ using the student model $S_{\text{\O}}$, (4) the diversity of both instructions and responses separately:
\begin{itemize}[leftmargin=*]
    \item \textbf{Response Quality} : We measure the quality of $R_i$ given $I_i$.
    First, we use \textbf{LLM-as-a-Judge} where we prompt an LM to return a discrete score between 1 and 5 that represents the quality of $R_i$. We employ two LM judges: (1) Prometheus-2-8x7B~\citep{kim2024prometheus}, an open-source LM specialized on assessing LM output and (2) GPT-4o, a proprietary LM widely used as a judge. We use different score rubrics for each domain, listed in Appendix~\ref{appendix:prompt_intrinsic_eval}. Next, we use \textbf{Reward Models} that predicts a scalar value score that represents the quality of $R_i$. We use Skywork-Reward-Llama-3.1-8B~\citep{liu2024skywork}, one of the top performing reward models on Reward Bench~\citep{lambert2024rewardbench}.
    
    \item \textbf{Instruction Complexity (LLM-as-a-Judge Score)}: We measure the difficulty of $I_i$ by prompting an LM to return a discrete score between 1 and 5 that represents the complexity of $I_i$. Similarly to evaluating response quality, we use Prometheus-2-8x7B and GPT-4o as a judge. The score rubric differs compared to that for evaluating response quality and we use a different score rubrics for each domain, listed in Appendix~\ref{appendix:prompt_intrinsic_eval}.
    \item \textbf{Perplexity of Response}: We measure the \textbf{perplexity} of $R_i$ conditioned on $I_i$ using the base model ${S}_{\text{\O}}$ (Llama-3.1-8B).
    \item \textbf{Instance Diversity}: We separately measure the average \textbf{cosine similarity} of instructions within $D_I = \{(I_i) \mid i = 1, \dots, n\}$ and responses within $D_R = \{(R_i) \mid i = 1, \dots, n\}$. This represents the extent to which each instruction or response is widely distributed (\textit{i.e.}, diverse)~\citep{ni2024mixeval}. We use \texttt{dunzhang/stella\_en\_400M\_v5}, a model that is both high-performing on the MTEB benchmark~\citep{muennighoff2023mteb} and efficient.
\end{itemize}

Due to page limits, the full results of the intrinsic evaluation are further detailed in Appendix~\ref{appendix:intrinsic_eval}.

\begin{figure}[!t] 
\centering
\includegraphics[width=0.45\textwidth]{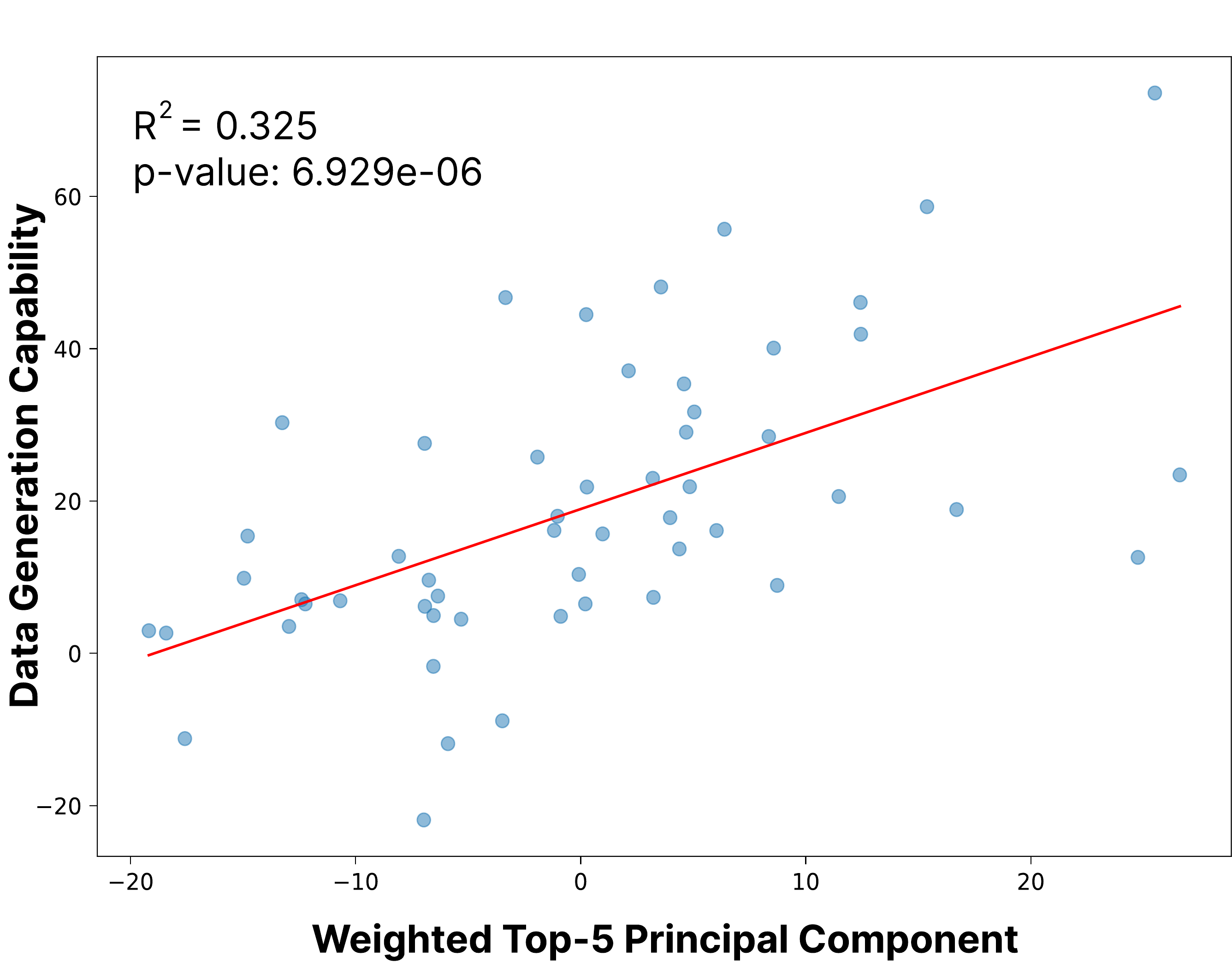}
\caption{\textbf{Principal Components from Intrinsic Metrics Show Stronger Correlation with Data Generation ability:} Linear regression using the weighted top-5 principal components yields a higher explained variance ($R^2$ = 0.325) and statistical significance ($p < 0.001$) compared to using problem-solving ability scores alone ($R^2 < 0.1$ or $p > 0.05$; see Figure~\ref{fig:linear_regression1}).}
\vspace{-3mm}
\label{fig:linear_regression2}
\end{figure}

\paragraph{Experiments} Inspired by the experiments from \citet{ruan2024observational}, we conduct a Principal Component Analysis (PCA) to investigate whether intrinsic evaluation metrics can explain the variability in \textsc{AgoraBench} results. We opt for PCA rather than multivariate linear regression due to the interdependence among our intrinsic evaluation metrics.

\begin{table}[!t]
\centering
\resizebox{\linewidth}{!}{
\begin{tabular}{lcc}
\toprule
\textbf{Intrinsic Metric} & \textbf{Loading Strength} & \textbf{Contribution}\\
\midrule
Prometheus Score (R.Q.)       &        0.256 & 12.18\%\\
Response Perplexity           &    0.252 & 12.00\%\\
GPT-4o Score (R.Q.)     &    0.246 & 11.71\%\\
Problem-solving Ability    &          0.240 & 11.42\%\\
Skywork-RM Score (R.Q.)      &       0.239 & 11.38\%\\
Prometheus Score (I.D.)  &            0.230 & 10.95\%\\
Diversity (I.D.)         &    0.226 & 10.76\%\\
GPT-4o Score (I.D.)    &          0.223 & 10.61\%\\
Diversity (R.Q.)    &     0.189 & 9.00\%\\
\bottomrule
\end{tabular}
}
\caption{\textbf{Mean Contributions of Intrinsic Metrics to Principal Components:} Each \textbf{loading strength} represents the average magnitude of a feature's loadings across all principal components and the \textbf{contribution} are normalized values to represent the relative percentage of each feature's loading strength in the overall component structure. `I.D.' refers to instruction difficulty metrics and `R.Q.' refers to response quality metrics.}
\label{table:contribution}
\end{table}

The results, shown in Figure~\ref{fig:pca}, reveal that the top five principal components explain approximately 93.4\% of the variance in \textsc{AgoraBench} results (39.2\%, 30.4\%, 11.9\%, 7.0\%, and 4.9\% respectively). Moreover, we find that analysis of the component weights reveals interpretable patterns. The first principal component (PC-1) is strongly influenced by instruction difficulty and diversity-related metrics. The second component (PC-2) is affected by response quality and instruction difficulty, while the third component (PC-3) combines diversity-related metrics, response quality, and the LM's problem-solving ability. 

Additionally, as shown in Table~\ref{table:contribution}, when we analyze the average loading strengths of each intrinsic evaluation metric (average magnitude of a feature's loadings across all principal components, indicating how strongly each metric influences the overall variance in the data), we observe that the contributions range from 0.189 to 0.256, indicating that all the intrinsic evaluation metrics contribute similarly to the PGR results. Also, we find that response quality-related metrics shows slightly stronger contributions than diversity-related metrics or instruction difficulty-related metrics to the PGR results.

Lastly, we predict data generation capabilities by performing linear regression on the top-5 principal components, weighting each component by its corresponding regression coefficient, as shown in Figure~\ref{fig:linear_regression2}. Compared to using problem-solving scores alone (Figure~\ref{fig:linear_regression1}), this approach yields a statistically significant relationship ($p < 0.001$) with improved explanatory power ($R^2 = 0.325$). However, the moderate $R^2$ value suggests that additional intrinsic metrics beyond our current set might be needed to better predict data generation capabilities.


\section{Conclusion}


%
In this paper, we introduce \textsc{AgoraBench}, a benchmark that systematically evaluates LMs' data generation capabilities through standardized settings and metrics. Looking ahead, we envision \textsc{AgoraBench} enabling two key advances in the field. First, since our findings suggest that problem-solving ability is not the primary determinant of data generation quality, researchers can use our benchmark to identify the core capabilities that make an effective data generator and potentially develop specialized LMs specialized in data generation. Second, \textsc{AgoraBench} can serve as a practical evaluation framework for practitioners to assess and improve their data generation pipelines - they can use their custom data generation methods, seed datasets, or meta-prompts and compare against our baseline settings. Through these complementary research and applied directions, \textsc{AgoraBench} aims to accelerate both our theoretical understanding of language models as data generators and their practical deployment in real-world applications.

\section*{Limitations \& Potential Risks}
Due to compute constraints and expensive API costs, the experimental setting of \textsc{AgoraBench} does not cover the scenarios where (1) different base models beyond Llama-3.1-8B are used and (2) whether our findings will hold when generating more than 10K instances.

Regarding the first limitation, while Llama-3.1-8B is widely adopted as a base model in the community, making our findings particularly relevant for researchers working with similar architectures, investigating the generalizability of our results across different base models remains an important direction for future research. Of particular interest is the variation in response perplexity—one of our intrinsic evaluation metric—across different model architectures, as this metric's behavior is inherently dependent on the base model.

For the second limitation, we conducted preliminary scaling experiments with more cost-effective language models (GPT-4o-mini, Llama-3.1-70B-Instruct, and Llama-3.1-8B-Instruct) generating up to 50K instances, as detailed in Appendix~\ref{sec:ablation}. Future work could explore whether the relative effectiveness rankings among different data generators remain consistent at larger scales, and investigate potential scaling laws in relation to our proposed Performance Gap Recovered metric. Such analysis could provide valuable insights into the optimal volume of synthetic data required for specific applications and model architectures.

Synthetic data holds significant potential in aligning LMs towards human preferences and controlling their behavior. While our study primarily focuses on improvements in mathematical reasoning, coding capabilities, and instruction following abilities, the broader implications for model alignment and control warrant careful consideration. Future work could explore these aspects more explicitly, particularly investigating how synthetic data generation techniques might influence model behavior beyond task performance, including potential biases, safety considerations, and the robustness of aligned behaviors across different contexts and applications. Lastly, using a proprietary model as a data generator raises concerns about potential biases, lack of transparency in the generation process, and the need to ensure proper permissions for its use. These factors go beyond performance and could be critical for practitioners to consider when choosing a data generation approach. Addressing such issues is important to ensure ethical and responsible usage.

\section*{Acknowledgements}

This research was supported in part by a gift from NEC Laboratories Europe. We thank the members of Neulab and L3Lab at CMU for helpful discussions. SW thanks Convergent Research.

\bibliography{acl2025}
\bibliographystyle{acl2025}

\newpage
\appendix

\clearpage

\section{Related Work}



Conventionally, training LMs on human-crafted data was considered the de facto standard for improving an LM's performance on downstream tasks~\citep{mishra2021cross,wei2021finetuned,wang2022super,longpre2023flan}. Yet, based on the in-context learning abilities of LMs~\citep{brown2020language}, a series of works demonstrated that LMs could generate novel instances that could be used as post-training data~\citep{wang2023self,honovich2022instruction,kim2023cot}.

Since then, different works have proposed various data generation methods and prompts to acquire high-quality data, using stronger LMs as data generators. For instance, \citet{alpaca} used the same data generation method as \citet{wang2023self}, but employed InstructGPT instead of GPT-3-Davinci and trained Llama-1 instead of T5. \citet{xu2024wizardlm} used ChatGPT as their data generator and proposed a method called Evol-Instruct that prompts the data generator to make an existing problem more complex than the original. \citet{mukherjee2023orca} used GPT-4 to generate data and improved the original response by adding a chain-of-thought explanation of how the answer was derived. \citet{xu2024magpie} proposed Magpie, a data generation method that first prompts an LM with an empty chat template to extract instructions, then iteratively prompts it to generate corresponding responses.

While developing new data generation methods is important, choosing which LM to use as a data generator is an equally crucial problem for both researchers and practitioners. To the best of our knowledge, \citet{xu2024stronger}, a contemporary work with our work, was the first attempt to measure various LMs' data generation capabilities using existing data generation methods. Yet, their settings were confined to our `response generation' method, whereas we also tested instance generation and quality enhancement methods.

\begin{figure*}[h] 
\centering
\includegraphics[width=\textwidth]{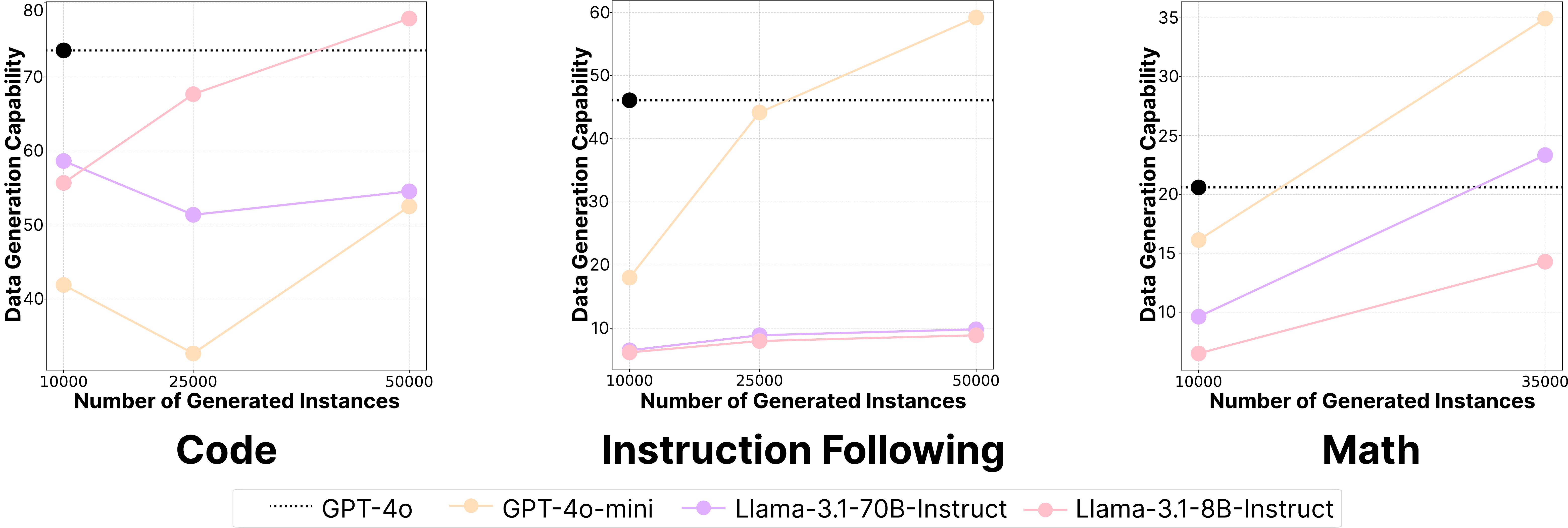}
\caption{\textbf{With a fixed budget, generating large amounts of data with weaker LMs could sometimes be more effective and cheaper than generating a few instances with stronger LMs:} Since GPT-4o-mini is 17 times cheaper than GPT-4o, generating 50K instances is 3.4 times cheaper than generating 10K instances with GPT-4o. Yet, generating 50K instances with GPT-4o-mini achieves higher PGR in instruction following and math domains compared to generating 10K instances with GPT-4o. }
\vspace{-3mm}
\label{fig:quantity_quality}
\end{figure*}

\begin{table*}[!t]
\caption{\textbf{Performance Gap Recovered (\%) results with different meta-prompts on instance generation and quality enhancement}. Llama models are instruction-tuned versions and that `Inst.' denotes instruction-following.}
\vspace{3mm}
\centering
\resizebox{\linewidth}{!}{
\begin{tabular}{lcccccccccccc}
\toprule
\multicolumn{1}{c}{\multirow{2}{*}{\textbf{Data Generator}}} & \multicolumn{4}{c}{\textbf{\textsc{AgoraBench} Meta-prompt}} &\multicolumn{4}{c}{\textbf{Unoptimized Meta-prompt}} &\multicolumn{4}{c}{\textbf{JSON-format Meta-prompt}} \\
\cmidrule(lr){2-5} \cmidrule(lr){6-9} \cmidrule(lr){10-13}
              & Math & Code & Inst. & Avg & Math & Code & Inst. & Avg & Math & Code & Inst. & Avg \\
\midrule
\multicolumn{13}{c}{\textbf{Instance Generation}} \\
\midrule
GPT-4o-mini & 16.1 & 41.9 & 18.0 & 25.3 & 12.4 & 36.8 & 17.6 & 22.3 & 13.8 & 20.5 & 19.5 & 17.9\\
Llama-3.1-70B  & 9.6 & 58.7 & 6.5 & 24.9 & 7.0 & 46.8 & 5.8 & 19.9 & 8.7 & 33.5 & 6.1 & 16.1\\
Llama-3.1-8B  & 6.5 & 55.7 & 6.2 & 22.8 & 0.7 & 43.6 & 4.5 & 16.3 & 6.7 & 31.4 & 4.4 & 14.2\\
\midrule
\multicolumn{13}{c}{\textbf{Quality Enhancement}} \\
\midrule
GPT-4o-mini & 17.8 & -11.2 & 9.9 & 5.5 & 13.0 & -6.3 & 9.4 & 5.4 & 15.4 & -13.0 & 9.2 & 3.8\\
Llama-3.1-70B  & -21.8 & 6.9 & 2.7 & -4.1 & -20.5 & -5.5 & 2.3 & -7.9 & -18.3 & 6.5 & 2.4 & -3.1\\
Llama-3.1-8B  & -1.7 & 15.4 & 3.0 & 5.6 & -6.6 & 3.7 & 3.5 & 0.2 & -2.7 & 12.0 & 3.9 & 4.4\\
\bottomrule
\end{tabular}
}
\label{table:ablation_results}
\end{table*}

\section{Further Analysis Experiments}\label{sec:ablation}


In this section, we further examine, two critical questions regarding data generation: (1) Should we prioritize quantity using cheaper LMs, or quality using more expensive ones? (Section~\ref{subsec:less_or_more}) And (2) What is the impact of meta-prompt design, particularly when comparing structured JSON format generation against traditional free-form approaches? (Section~\ref{subsec:meta_prompt}). 

\subsection{Quantity or quality?}
\label{subsec:less_or_more}

In Section~\ref{sec:extrinsic_eval}, we demonstrated that in some cases, cheaper LMs can be more effective data generators than their expensive counterparts when producing a fixed number of instances, though expensive models generally perform better. This raises a practical question: Is it more effective to generate a larger quantity of instances using cheaper models rather than fewer instances with more expensive ones?

We scale up our experiment to generate up to 50K instances using GPT-4o-mini, Llama-3.1-70B-Instruct, and Llama-3.1-8B-Instruct across three domains in the instance generation scenario. As shown in Figure~\ref{fig:quantity_quality}, generating 50K instances with GPT-4o-mini resulted in better performance than generating 10K instances with GPT-4o at instruction following and math domains and Llama-3.1-8B-Instruct showed similar patterns in code domain. Given that these LMs are at least five times more cost-effective than GPT-4o, our findings suggest that generating larger volumes of synthetic data with more affordable LMs may be more advantageous than generating smaller datasets with expensive ones. Furthermore, this suggests that instruction diversity or response diversity could affect the PGR results when comparing two settings with different number of training instances.

\subsection{Effect of Meta-prompts}\label{subsec:meta_prompt}

Recently, \citet{tam2024let} has shown that LMs' problem-solving abilities decrease when generating responses in structured formats (e.g., JSON). Given practitioners' preference for structured outputs when using LMs~\citep{shorten2024structuredrag,liang2024universal} it's important to investigate whether this format affects data generation performance. Additionally, we examine the impact of meta-prompt design on generation quality.

To investigate these questions, we create four additional meta-prompts for comparison. For each setting (instance generation and quality enhancement), we had two co-authors create meta-prompts: one developed an unoptimized version (spending less than 10 minutes)\footnote{This contrasts with the main experiments' meta-prompts, which were developed over 2+ hours through iterative trial-and-error during the initial experimental phase.}, while the other created a JSON-format version.

Table~\ref{table:ablation_results} presents our findings. Compared to the other meta-prompts, the \textsc{AgoraBench} meta-prompt achieves the highest scores in five out of six settings, demonstrating the robustness of the setting in \textsc{AgoraBench}. Comparing the \textsc{AgoraBench} meta-prompts with unoptimized versions reveals a 3.97\% performance gap on average, highlighting the importance of meta-prompt optimization. Furthermore, \textsc{AgoraBench} meta-prompts using free-form generation achieve 4.45\% higher performance compared to JSON-format prompts. This aligns with recent findings that structured format requirements may compromise LM output quality~\citep{tam2024let}.

\section{Response Generation Seed Dataset Construction}\label{appendix:webinstruct}

In \textsc{AgoraBench}, we prepare seed datasets for each domain (instruction-following, math, code) separately in order to prevent positive or negative transfer that occurs during training, which could make it difficult to ground the PGR results to the quality of the synthetic data and LMs' data generation capabilities.

We use the WebInstruct data~\citep{yue2024mammoth2} for math and instruction-following domains in quality enhancement settings. However, the WebInstruct data does not provide labels of whether the given instance is a math problem or not. Hence, we prompted GPT-4o-mini-2024-07-18 to classify it using the following prompt:

\begin{mybox}{Domain classification for seed data construction}
Classify whether the following ''Instance`` consisted of an ''Instruction`` and ''Response`` is either related to:

1. Math-related task such as requiring an answer to a problem, proving a theorem or explaining about a mathematical concept.

2. Other tasks
\\

Provide your answer in only either ''1`` or ''2``, without any greeting message or comment.
\\

\# Instance:

Instruction: <input>

Response: <output>
\\

\# Decision:
\end{mybox}

\section{Problem Solving Abilities of LMs evaluated as Data Generators}\label{appendix:teacher_model_performance}

The evaluation results of GPT-4o-2024-08-06, GPT-4o-mini-2024-07-18, Claude-3.5-Sonnet-2024-06-20, Llama-3.1-405B-Instruct, Llama-3.1-70B-Instruct, and Llama-3.1-8B-Instruct are listed in Table~\ref{table:teacher_model_perf}. We use the settings listed in Appendix~\ref{appendix:hyper_params}.

\begin{table}[!t]
\centering
\fontsize{8}{10}\selectfont
\begin{tabular}{c|c}
\toprule
\multicolumn{2}{c}{\multirow{1}{*}{Inference Hyper-parameter}}\\
\midrule
\textbf{Temperature} & 0.2 (math) \& 0.0 (other domains)\\
\textbf{Top\_p} &  0.95\\
\textbf{Max New Tokens} & 1024 \\ 
\textbf{Repetition Penalty} & 1.03 \\ 
\midrule
\multicolumn{2}{c}{\multirow{1}{*}{Training Hyper-parameter}}\\
\midrule
\textbf{Base Model} & meta-llama/Llama-3.1-8B\\
\textbf{Torch dtype} &  bfloat16\\
\textbf{Epoch} & 5 \\ 
\textbf{Max Seq Length} & 4096\\
\textbf{Learning Rate} & 1e-5\\
\textbf{Train Batch Size} & 4\\
\textbf{Gradient Accumulation} & 8\\
\textbf{GPU} & H100 (80GB) x 4\\
\textbf{Random Seed} & 42\\
\textbf{Training Method} & Supervised Fine-tuning\\
\bottomrule
\end{tabular}%
\caption{\footnotesize Hyper-parameters used for inference.}
\label{tab:hyperparameter}
\end{table}

\section{Details for Training and Evaluating Student Models}\label{appendix:hyper_params}

The hyper-parameters used for training student models and hyper-parameters used for evaluating both student models and LMs employed as the data generator are listed in Table~\ref{tab:hyperparameter}.

For evaluation on MBPP and HumanEval, we use the Evalplus library~\citep{evalplus}. For evaluation on AlpacaEval and ArenaHard, we use the official library, respectively~\citep{dubois2024length,li2024crowdsourced}. For GSM8K and MATH, we use the datasets provided in huggingface and use our manual script. All the evaluation scripts are publicly available at our repository.

\section{Intrinsic Evaluation of \textsc{AgoraBench}}\label{appendix:intrinsic_eval}
The intrinsic evaluation results are listed in Table~\ref{table:intrinsic_results}.

\section{\textsc{AgoraBench} Meta-prompts}

Due to space limits, we present the meta-prompts in our repository and the following \href{https://drive.google.com/drive/folders/1EfYwgo0T7tJGDnpQ2CUZGciPBfucu73g?usp=sharing}{link}.

\section{Prompt for Intrinsic Evaluation}\label{appendix:prompt_intrinsic_eval}

In the following pages, we list the prompt used for assessing response quality and instruction difficulty with GPT-4o and Prometheus-2-8x7B as well as the score rubrics for used for each domain (instruction-following, math, code).

\begin{mybox}{Evaluation Prompt for Response Quality and Instruction Difficulty}
\#\#\#Task Description:

An instruction (might include an Input inside it), a response to evaluate, and a score rubric representing a evaluation criteria are given.

1. Write a detailed feedback that assess the quality of the response strictly based on the given score rubric, not evaluating in general.

2. After writing a feedback, write a score that is an integer between 1 and 5. You should refer to the score rubric.

3. The output format should look as follows: "Feedback: (write a feedback for criteria) [RESULT] (an integer number between 1 and 5)"

4. Please do not generate any other opening, closing, and explanations.

\#\#\#The instruction to evaluate:

\{instruction\}

\#\#\#Response to evaluate:

\{response\}

\#\#\#Score Rubric:
\{score\_rubric\}

\#\#\#Feedback: 
\end{mybox}

\clearpage

\begin{mybox}{Math Response Quality Score Rubirc}
Does the solution demonstrate mathematical correctness, reasoning, clarity, and precision?

Score 1 Description: The solution is incorrect or mathematically flawed, with major errors in reasoning, calculations, or logic, making the answer unusable.

Score 2 Description: The solution contains relevant or partially correct information, but has significant errors in calculations or reasoning that substantially affect the result.

Score 3 Description: The solution is mostly correct but may contain minor mistakes or gaps in reasoning. The overall structure and approach are sound, but some calculations or logic may need refinement.

Score 4 Description: The solution is correct, well-reasoned, and clear, though there may be slight room for improvement or minor refinements to become a perfect solution to the problem.

Score 5 Description: The solution is excellent, fully correct, and demonstrates a high level of mathematical precision, clarity, and creativity, with well-articulated reasoning and no errors.
\end{mybox}

\begin{mybox}{Instruction-following Response Quality Score Rubric}
Does the response consider a wide range of factors such as the helpfulness, relevance, accuracy, depth, creativity, and level of detail?
\\

Score 1 Description: The response is not helpful at all or seems helpful on the surface but is actually incorrect such as including incorrect information, naive miscalculations, or unexecutable code.

Score 2 Description: The response contains some relevant or helpful information, but also has major flaws interms of factuality, accuracy, and relevance.

Score 3 Description: The response is mostly correct but minor flaws regarding factuality, accuracy, and relevance still exists, while it is overall an okay response.

Score 4 Description: The response is accurate, relevant, and helpful, although there are some slight improvements that could be made when an expert evaluates the response.

Score 5 Description: The response is excellent. It is completely factual, accurate, relevant, and helpful, demonstrating a high degree of depth and creativity.
\end{mybox}

\begin{mybox}{Code Response Quality Score Rubric}
How effective, efficient, and logically sound is the code solution, focusing on performance, executability, and correctness?
\\

Score 1 Description: The code contains fundamental logic or syntax errors, making it incorrect or unexecutable. It fails to complete the intended task or produces entirely incorrect outputs.

Score 2 Description: The code is partially functional but contains major logic errors or inefficiencies that significantly impact performance or correctness. It may run but produces incorrect or incomplete results.

Score 3 Description: The code is mostly correct and executable, though there may be minor logic issues, inefficiencies, or suboptimal use of data structures or algorithms. The solution functions as intended, but improvements could make it more efficient or robust.

Score 4 Description: The code is fully correct, functional, and reasonably efficient. It completes the task as intended, balancing performance with logical soundness. Minor optimizations could still enhance performance.

Score 5 Description: The code is fully correct, optimally efficient, and logically robust, providing the best possible performance for the task. It executes flawlessly without errors or any significant room for improvement.
\end{mybox}

\begin{mybox}{Instruction-following Instruction Difficulty Score Rubric}
How complex and challenging is the given instruction to answer perfectly?
\\

Score 1 Description: The instruction requires only factual knowledge, without any need for reasoning or critical thinking. A straightforward, single-step response suffices.

Score 2 Description: The instruction requires some reasoning, such as explaining a concept involving multiple simple ideas, solving a straightforward problem, or providing a response that involves a few logical steps, though still simple in nature.

Score 3 Description: The instruction requires a substantial amount of reasoning and the integration of multiple related concepts. Answering it accurately involves a multi-step process and may require intermediate-level knowledge or analytical thinking.

Score 4 Description: The instruction requires advanced reasoning, demanding deep understanding of complex concepts or substantial problem-solving. Answering it requires carefully navigating multiple interrelated ideas or steps, often involving specialized knowledge or sophisticated analytical skills.

Score 5 Description: The instruction is exceptionally challenging and requires high-level reasoning or novel problem-solving. It involves extensive conceptual understanding, abstraction, and potentially innovative thinking, with substantial effort required to arrive at an accurate and complete answer.
\end{mybox}

\begin{mybox}{Math Instruction Difficulty Score Rubric}
How complex and challenging is the math problem to solve?
\\

Score 1 Description: The problem requires only simple operations or direct application of a single, basic concept. Minimal reasoning is needed, and the solution follows immediately from applying a known rule or formula.

Score 2 Description: The problem requires basic reasoning and involves applying a familiar formula or concept with slight variation. It may involve a straightforward multi-step process, but each step is clear and relies on commonly used methods.

Score 3 Description: The problem requires moderate reasoning, combining multiple concepts that interact in a meaningful way. Solving it involves several steps and may require logical sequencing or some abstraction, but the approach is approachable with a solid foundational understanding.

Score 4 Description: The problem demands advanced reasoning, involving multiple interdependent concepts that require careful coordination. Solution steps are less obvious, requiring critical thinking and possibly choosing between multiple solution paths. Solving the problem involves more abstract reasoning or creative application of concepts.

Score 5 Description: The problem is extremely complex and demands sophisticated reasoning and problem-solving skills. It may involve novel combinations of concepts, intricate logical chains, or innovative approaches to solve. This level typically requires significant abstraction, exploration of unconventional methods, and flexibility in adapting mathematical tools.
\end{mybox}

\begin{mybox}{Code Instruction Difficulty Score Rubric}
How complex and challenging is the coding problem to solve?
\\

Score 1 Description: The problem involves implementing simple functionality or a direct operation. It requires minimal logic, with a straightforward approach and no complex decision-making.

Score 2 Description: The problem requires basic control flow, such as using loops or conditional statements. The logic is clear and sequential, with minimal interaction between different parts of the code.

Score 3 Description: The problem involves intermediate logic, combining multiple programming constructs and requiring a coherent structure. Solving it requires handling a sequence of steps with basic data manipulation, but follows a familiar, manageable approach.

Score 4 Description: The problem demands advanced reasoning and use of complex data structures or algorithms. It involves non-trivial interactions, such as managing multiple components and optimizing for efficiency. The solution requires significant algorithmic thinking and structured problem decomposition.

Score 5 Description: The problem is extremely complex, requiring sophisticated algorithm design, efficient data handling, and advanced techniques. It demands innovative approaches, with intricate component interactions and constraints that need careful optimization. Solving it typically requires deep problem-solving skills and adaptability across programming paradigms.
\end{mybox}

\begin{table*}[!t]
\caption{Problem-solving abilities of LMs measured by benchmark scores.}
\centering
\resizebox{0.7\linewidth}{!}{
\begin{tabular}{lccccccc}
\toprule
\multicolumn{1}{c}{\multirow{3}{*}{\textbf{Data Generator}}}& \multicolumn{7}{c}{\textbf{Problem-solving ability}}\\
\cmidrule{2-8}
             & \multicolumn{1}{c}{\multirow{2}{*}{\textbf{GSM8K}}} & \multicolumn{1}{c}{\multirow{2}{*}{\textbf{MATH}}} & \multicolumn{1}{c}{\multirow{2}{*}{\textbf{MBPP}}} & \textbf{Human} & \textbf{Alpaca} & \textbf{Arena} & \multicolumn{1}{c}{\multirow{2}{*}{\textbf{Average}}}\\
              &&&&\textbf{Eval}&\textbf{Eval 2.0} & \textbf{Hard} & \\
\cmidrule(lr){1-1} \cmidrule(lr){2-2} \cmidrule(lr){3-3} \cmidrule(lr){4-4} \cmidrule(lr){5-5} \cmidrule(lr){6-6} \cmidrule(lr){7-7} \cmidrule(lr){8-8}
GPT-4o & 96.1 & 76.6 & 86.2 & 91.5 & 57.5 & 77.9 & 80.9\\
GPT-4o-mini &  93.2 & 70.2 & 85.7 & 88.4 & 50.7 & 64.2 & 75.4\\
Claude-3.5-Sonnet &96.4 & 71.1 & 89.2 & 92.0 & 52.4 & 82.0 & 80.5\\
Llama-3.1-405B & 96.8 & 73.8 & 84.5 & 89.0 & 39.3 & 66.8 & 75.0\\
Llama-3.1-70B & 95.1 & 68.0 & 84.2 & 80.5 & 38.1 & 51.6 & 69.6 \\
Llama-3.1-8B  & 78.9 & 34.6 & 68.5 & 69.5 & 24.2 & 25.5 & 50.2 \\
\bottomrule
\end{tabular}
}
\label{table:teacher_model_perf}
\end{table*}

\begin{table*}[!t]
\caption{Intrinsic evaluation results of \textsc{AgoraBench}.}
\vspace{3mm}
\centering
\resizebox{0.9\linewidth}{!}{
\begin{tabular}{lcccccccccccc}
\toprule
\multicolumn{1}{c}{\multirow{2}{*}{\textbf{Data Generator}}} & \multicolumn{4}{c}{\textbf{Instance Generation}} &\multicolumn{4}{c}{\textbf{Response Generation}} &\multicolumn{4}{c}{\textbf{Quality Enhancement}} \\
\cmidrule(lr){2-5} \cmidrule(lr){6-9} \cmidrule(lr){10-13}
              & Math & Code & Inst. Follow & Avg & Math & Code & Inst. Follow & Avg & Math & Code & Inst. Follow & Avg \\
\midrule
\multicolumn{13}{c}{\multirow{1}{*}{\textbf{Instruction Difficulty (LLM-as-a-Judge; GPT-4o Score)}}}\\
\midrule
GPT-4o (2024-08-06) & 2.92 & 3.48 & 3.06 & 3.16 & 2.27 & 2.21 & 1.41 & 1.97 & 2.44 & 1.51 & 1.79 & 1.91\\
GPT-4o-mini (2024-07-18) & 2.38 & 3.42 & 2.89 & 2.90 & 2.27 & 2.21 & 1.41 & 1.97 & 2.47 & 1.38 & 1.81 & 1.89\\
Claude-3.5-Sonnet (2024-06-20) & 3.24 & 4.03 & 3.54 & 3.60 & 2.27 & 2.21 & 1.41 & 1.97 & 2.47 & 1.52 & 1.83 & 1.94\\
Llama-3.1-405B-Instruct  & 2.74 & 3.50 & 2.87 & 3.04 & 2.27 & 2.21 & 1.41 & 1.97 & 2.45 & 1.47 & 1.85 & 1.92\\
Llama-3.1-70B-Instruct  & 2.87 & 3.45 & 2.96 & 3.09 & 2.27 & 2.21 & 1.41 & 1.97 & 2.48 & 1.49 & 1.87 & 1.95 \\
Llama-3.1-8B-Instruct  & 3.00 & 3.52 & 3.08 & 3.20 & 2.27 & 2.21 & 1.41 & 1.97 & 2.43 & 1.49 & 1.83 & 1.92\\
\midrule
\multicolumn{13}{c}{\multirow{1}{*}{\textbf{Instruction Difficulty (LLM-as-a-Judge; Prometheus-2-8x7B Score)}}}\\
\midrule
GPT-4o (2024-08-06) & 3.73 & 3.57 & 3.95 & 3.75 & 3.00 & 2.76 & 2.24 & 2.67 & 3.37 & 2.14 & 2.50 & 2.67\\
GPT-4o-mini (2024-07-18) & 3.44 & 3.38 & 3.94 & 3.59 & 3.00 & 2.76 & 2.24 & 2.67 & 3.36 & 1.98 & 2.53 & 2.63\\
Claude-3.5-Sonnet (2024-06-20) & 4.11 & 4.51 & 4.45 & 4.36 & 3.00 & 2.76 & 2.24 & 2.67 & 3.38 & 2.24 & 2.61 & 2.74\\
Llama-3.1-405B-Instruct  & 3.63 & 3.27 & 3.84 & 3.58 & 3.00 & 2.76 & 2.24 & 2.67 & 3.35 & 2.11 & 2.64 & 2.70\\
Llama-3.1-70B-Instruct & 3.72 & 3.43 & 3.94 & 3.69 & 3.00 & 2.76 & 2.24 & 2.67 & 3.32 & 2.21 & 2.76 & 2.76\\
Llama-3.1-8B-Instruct  & 3.86 & 3.48 & 3.99 & 3.78 & 3.00 & 2.76 & 2.24 & 2.67 & 3.30 & 2.09 & 2.67 & 2.68\\
\midrule
\multicolumn{13}{c}{\multirow{1}{*}{\textbf{Instruction Difficulty (Perplexity)}}}\\
\midrule
GPT-4o (2024-08-06) & 2.13 & 1.28 & 3.44 & 2.28 & 2.26 & 4.23 & 3.41 & 3.30 & 2.03 & 3.60 & 3.83 & 3.15\\
GPT-4o-mini (2024-07-18) & 2.05 & 1.31 & 3.32 & 2.23 & 2.28 & 2.12 & 3.20 & 2.53 & 2.08 & 5.50 & 3.97 & 3.85 \\
Claude-3.5-Sonnet (2024-06-20) & 2.04 & 1.34 & 3.18 & 2.19 & 2.16 & 3.48 & 3.63 & 3.09 & 1.99 & 2.46 & 3.04 & 2.50\\
Llama-3.1-405B-Instruct  & 1.96 & 1.29 & 2.19 & 1.81 & 1.90 & 1.91 & 2.42 & 2.08 & 2.10 & 3.10 & 3.90 & 3.03\\
Llama-3.1-70B-Instruct  & 1.78 & 1.27 & 2.19 & 1.74 & 1.86 & 1.72 & 2.52 & 2.03 & 2.12 & 2.84 & 3.98 & 2.98\\
Llama-3.1-8B-Instruct  & 1.83 & 1.33 & 2.08 & 1.74 & 1.98 & 1.81 & 2.48 & 2.09 & 2.06 & 3.17 & 3.98 & 3.07\\
\midrule
\multicolumn{13}{c}{\multirow{1}{*}{\textbf{Response Quality (LLM-as-a-Judge; GPT-4o Score)}}}\\
\midrule
GPT-4o (2024-08-06) & 3.72 & 3.95 & 4.42 & 4.03 & 3.99 & 3.79 & 4.44 & 4.07 & 3.62 & 3.66 & 3.99 & 3.76\\
GPT-4o-mini (2024-07-18) & 3.96 & 3.96 & 4.35 & 4.09 & 3.85 & 3.76 & 4.41 & 4.01 & 3.57 & 3.22 & 3.96 & 3.58\\
Claude-3.5-Sonnet (2024-06-20) & 3.39 & 4.03 & 4.34 & 3.92 & 3.80 & 3.75 & 4.24 & 3.93 & 3.64 & 3.77 & 4.29 & 3.90\\
Llama-3.1-405B-Instruct  & 3.20 & 3.74 & 4.13 & 3.69 & 3.51 & 3.76 & 4.29 & 3.85 & 3.36 & 3.37 & 3.80 & 3.51\\
Llama-3.1-70B-Instruct  & 2.97 & 3.59 & 4.12 & 3.56 & 3.31 & 3.65 & 4.22 & 3.72 & 3.23 & 3.22 & 3.80 & 3.42\\
Llama-3.1-8B-Instruct  & 1.99 & 2.51 & 3.82 & 2.77 & 2.90 & 3.26 & 4.17 & 3.44 & 3.05 & 2.76 & 3.52 & 3.11\\
\midrule
\multicolumn{13}{c}{\multirow{1}{*}{\textbf{Response Quality (LLM-as-a-Judge; Prometheus-2-8x7B Score)}}}\\
\midrule
GPT-4o (2024-08-06) & 3.93 & 3.49 & 4.07 & 3.83 & 4.02 & 3.28 & 3.97 & 3.76 & 3.98 & 3.28 & 3.69 & 3.65\\
GPT-4o-mini (2024-07-18) & 4.05 & 3.46 & 4.04 & 3.85 & 3.96 & 3.39 & 3.93 & 3.76 & 3.92 & 3.04 & 3.73 & 3.57\\
Claude-3.5-Sonnet (2024-06-20) & 3.95 & 3.37 & 4.04 & 3.78 & 3.94 & 3.29 & 3.83 & 3.69 & 4.00 & 3.48 & 4.03 & 3.84\\
Llama-3.1-405B-Instruct  & 3.76 & 3.24 & 3.92 & 3.64 & 3.81 & 3.42 & 3.91 & 3.71 & 3.78 & 3.23 & 3.66 & 3.56\\
Llama-3.1-70B-Instruct  & 3.68 & 3.36 & 3.91 & 3.65 & 3.73 & 3.37 & 3.86 & 3.65 & 3.73 & 3.20 & 3.62 & 3.52\\
Llama-3.1-8B-Instruct  & 3.22 & 3.06 & 3.81 & 3.36 & 3.62 & 3.24 & 3.88 & 3.58 & 3.68 & 3.08 & 3.49 & 3.42\\
\midrule
\multicolumn{13}{c}{\multirow{1}{*}{\textbf{Response Quality (Reward Model; Skywork-RM-8B Score)}}}\\
\midrule
GPT-4o (2024-08-06) & 13.90 & 23.20 & 27.79 & 21.63 & 8.82 & -0.10 & 8.60 & 5.77 & 4.86 & -7.48 & -4.73 & -2.45\\
GPT-4o-mini (2024-07-18) & 5.74 & -1.18 & 7.80 & 4.12 & 13.71 & 23.74 & 20.71 & 19.39 & 3.42 & -12.93 & -5.16 & -4.89\\
Claude-3.5-Sonnet (2024-06-20) & 10.67 & 20.22 & 18.20 & 16.36 & 5.56 & 1.67 & 0.17 & 2.46 & 6.29 & -5.31 & 10.76 & 3.92\\
Llama-3.1-405B-Instruct  & -0.50 & 19.54 & 13.63 & 10.89 & -1.23 & 2.68 & 10.65 & 4.03 & -1.89 & -10.43 & -7.35 & -6.56\\
Llama-3.1-70B-Instruct  & -2.17 & 20.42 & 11.22 & 9.83 & -3.26 & 2.17 & 5.85 & 1.59 & -3.04 & -11.54 & -8.60 & -7.72\\
Llama-3.1-8B-Instruct  & -7.71 & 7.16 & 3.45 & 0.97 & -3.89 & -1.53 & 8.72 & 1.10 & -3.68 & -12.61 & -10.15 & -8.81\\
\midrule
\multicolumn{13}{c}{\multirow{1}{*}{\textbf{Instruction Diversity (c-dist)}}}\\
\midrule
GPT-4o (2024-08-06) & 0.4170 & 0.4640 & 0.3047 & 0.3952 & 0.3737& 0.3958 & 0.3386 & 0.3694 & 0.4263 & 0.4870 & 0.2943& 0.4025\\
GPT-4o-mini (2024-07-18) & 0.4091 & 0.5127 & 0.3013 & 0.4077 & 0.3737 & 0.3958& 0.3386& 0.3694 & 0.4270 & 0.4670 & 0.2956 & 0.3965\\
Claude-3.5-Sonnet (2024-06-20) & 0.4124 & 0.4872 & 0.2940 & 0.3979& 0.3737 & 0.3958& 0.3386& 0.3694 & 0.4307 & 0.4903 & 0.2921 & 0.4044\\
Llama-3.1-405B-Instruct  & 0.3996 & 0.5411 & 0.2789 & 0.4065& 0.3737& 0.3958& 0.3386& 0.3694 & 0.4344 & 0.4796 & 0.3033 & 0.4058\\
Llama-3.1-70B-Instruct  & 0.4003 & 0.5015 & 0.2682 & 0.3900& 0.3737& 0.3958& 0.3386& 0.3694 & 0.4232 & 0.4756 & 0.3018 & 0.4022\\
Llama-3.1-8B-Instruct  & 0.4201 & 0.4785 & 0.2956 & 0.3981& 0.3737& 0.3958& 0.3386& 0.3694 & 0.4302 & 0.4619 & 0.2984 & 0.3968\\
\midrule
\multicolumn{13}{c}{\multirow{1}{*}{\textbf{Response Diversity (c-dist)}}}\\
\midrule
GPT-4o (2024-08-06) & 0.4564 & 0.5347 & 0.2918 & 0.4276 & 0.4126 & 0.4719 & 0.2714 & 0.3853 & 0.4455 & 0.5065 & 0.2271 & 0.3930\\
GPT-4o-mini (2024-07-18) & 0.4558 & 0.5719 & 0.2814 & 0.4364 & 0.4095 & 0.4726 & 0.2811 & 0.3877 & 0.4577 & 0.5184 & 0.2257 & 0.4006\\
Claude-3.5-Sonnet (2024-06-20) & 0.4719 & 0.5648 & 0.3220 & 0.4529 & 0.4156 & 0.4647 & 0.2788 & 0.3864 & 0.4610 & 0.5141 & 0.2325 & 0.4025\\
Llama-3.1-405B-Instruct  & 0.4490 & 0.6122 & 0.2523 & 0.4378 & 0.4037 & 0.4737 & 0.2551 & 0.3775 & 0.4633 & 0.5155 & 0.2239 & 0.4009\\
Llama-3.1-70B-Instruct  & 0.4520 & 0.5771 & 0.2596 & 0.4296 & 0.4012 & 0.4784 & 0.2530 & 0.3775 & 0.4571 & 0.5134 & 0.2233 & 0.3979\\
Llama-3.1-8B-Instruct  & 0.4768 & 0.5651 & 0.2660 & 0.4360 & 0.4077 & 0.4778&0.2530 & 0.3795 & 0.4738 & 0.5143 & 0.2254 & 0.4045\\
\bottomrule
\end{tabular}
}
\label{table:intrinsic_results}
\end{table*}

\end{document}